\documentclass[a4paper,conference]{IEEEtran}
\IEEEoverridecommandlockouts
\usepackage{cite}
\usepackage{amsmath,amssymb,amsfonts}
\usepackage{algorithmic}
\usepackage{graphicx}
\usepackage{xcolor}
\usepackage[misc]{ifsym} 
\def\BibTeX{{\rm B\kern-.05em{\sc i\kern-.025em b}\kern-.08em
    T\kern-.1667em\lower.7ex\hbox{E}\kern-.125emX}}
\usepackage{hyperref}    
\begin{document}
\title{DmifNet: 3D Shape Reconstruction Based on Dynamic Multi–Branch Information Fusion\\
% {\footnotesize \textsuperscript{*}Note: Sub-titles are not captured in Xplore and
% should not be used}
\thanks{
${^*}$ represents corresponding author. This work was supported in part by the National Science Foundation of China under Grant 61662059, and in part by the Natural Science Foundation of China under Grant 62062056.}
}

\author{\IEEEauthorblockN{1\textsuperscript{st} Lei Li}
\IEEEauthorblockA{\textit{School of Information Engineering} \\
\textit{Ningxia University}\\
YinChuan, China \\
lliicnxu@163.com}
\and
\IEEEauthorblockN{2\textsuperscript{nd} Suping Wu ${^*}$}
\IEEEauthorblockA{\textit{School of Information Engineering} \\
\textit{Ningxia University}\\
YinChuan, China\\
wspg123@163.com}

}
%  $^{(\textrm{\Letter})}$
\maketitle

\begin{abstract}
3D object reconstruction from a single-view image is a long-standing challenging problem. Previous work was difficult to accurately reconstruct 3D shapes with a complex topology which has rich details at the edges and corners. Moreover, previous works used synthetic data to train their network, but domain adaptation problems occurred when tested on real data. In this paper, we propose a Dynamic Multi-branch Information Fusion Network (DmifNet) which can recover a high-fidelity 3D shape of arbitrary topology from a 2D image. Specifically, we design several side branches from the intermediate layers to make the network produce more diverse representations to improve the generalization ability of network. In addition, we utilize DoG (Difference of Gaussians) to extract edge geometry and corners  information from input images. Then, we use a separate side branch network to process the extracted data to better capture edge geometry and corners feature information. Finally, we dynamically fuse the information of all branches to gain final predicted probability. Extensive qualitative and quantitative experiments on a large-scale publicly available dataset demonstrate the validity and efficiency of our method. Code and models are publicly available at \hyperlink{https://github.com/leilimaster/DmifNet}{https://github.com/leilimaster/DmifNet}.
\end{abstract}


\section{Introduction}
3D reconstruction is an elementary problem of image processing and computer vision thanks to many potential useful real-world applications such as robotics, autonomous driving and augmented reality. Recently, 3D reconstruction based on deep learning has achieved remarkable results in many directions, such as 3D shape reconstruction from single view or multiple views \cite{wu2016learning,choy20163d,tatarchenko2019single} and shape completion \cite{xu2019disn}. Generally, most of the 3D reconstruction work is based on CNN encoder-decoder architecture\cite{choy20163d ,richter2018matryoshka}. Specifically, single-view image reconstruction tasks usually take 2D CNN to encode 2D images and use different decoders to produce different final representations according to what representations the task needs. For example, if voxels\cite{wu2018learning ,zhang2018learning}are expected to be the final representation, 3D CNN will be selected as the decoder.

\begin{figure}[htbp]
\centerline{\includegraphics[width=8.5cm]{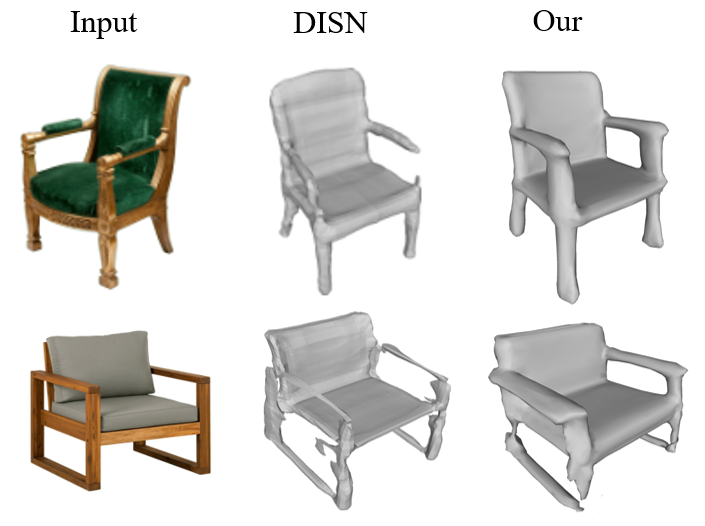}}
\caption{Single image reconstruction using a state-of-the-art method DISN\cite{xu2019disn}, and our method on real images.}
\label{fig}
\end{figure}

\begin{figure*}[htbp]
\centerline{\includegraphics[width=18cm]{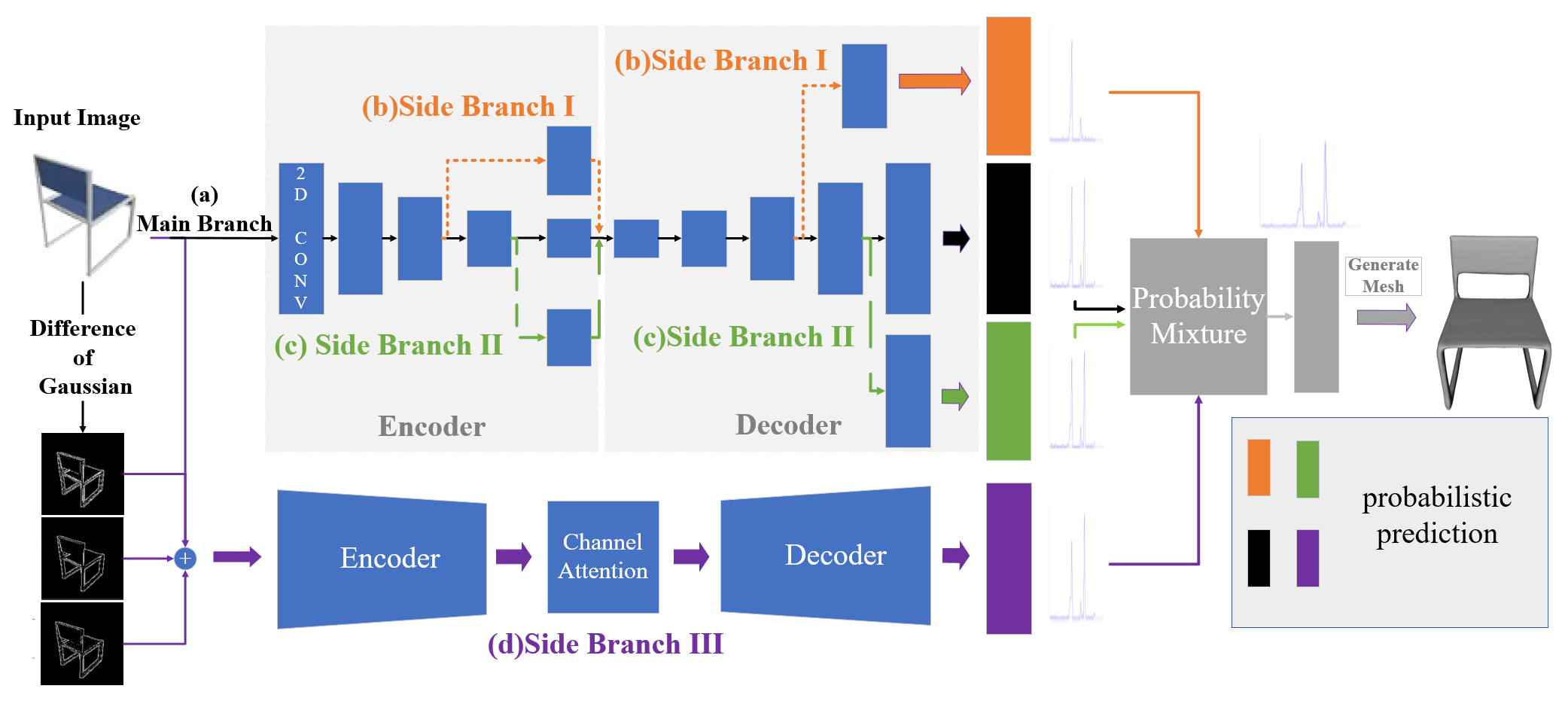}}
\caption{The workflow of the proposed DmifNet framework. Our model has a main branch and three side branches: (a) The main branch uses the autoencoder to process the sample data and get the prediction results. (b)(c)Branches I and II process data by exploiting sub-branches from different intermediate layers of the main branch. (d)Branch III first uses the DoG to process the samples to obtain the Gaussian difference map, then we concatenate the original 
input image and Gaussian difference map as input information to predict result. Finally, we dynamically fuse the prediction results of the main branch and the side branches to get final prediction results. Best viewed in color.
}
\label{fig}
\end{figure*}

Historically, single-view image 3D reconstruction was achieved via shape-from-shading \cite{zhang1999shape,horn1970shape}. These work inferred the depth information of the visible surface through multiple clues (e.g. texture, defocus) and structural information from a single image. Recently, according to the output representation, the existing work on learning-based 3D reconstruction can be categorized into mesh-based, point-based and voxel-based. Since there is no clear way to generate a valid mesh, mesh-based representation is facing great challenges. Then, Wang ${et}$ ${al.}$ \cite{wang2018pixel2mesh} used graph convolutional neural network \cite{scarselli2008graph} to make an ellipsoid template gradually become the target object, but the result was usually limited to the spherical topology.  Fan ${et}$ ${al.}$ \cite{fan2017point}introduced point clouds as an output representation for 3D reconstruction. However, the point-based representation requires many complex post-processing steps to generate a 3D mesh. Some work \cite{wu2016learning,choy20163d,wu20153d} have considered voxel-based representations and used 3D CNN to reconstruct the 3D shape from a single image. But, this is only possible with shallow architectures and small batch sizes due to memory limitation, which leads to slow training. To address the impact of memory limitation, Tatarchenko ${et}$ ${al.}$ \cite{tatarchenko2017octree} performed hierarchical partitioning of the output space for computational and storage efficiency, which helped predict higher resolution 3D shapes. Previous work chose to train their model on synthetic data with ground truth 3D information, but suffered from domain adaptation when tested on real data. To address the problem of domain adaptation, Wu ${et}$ ${al.}$ \cite{wu2017marrnet} proposed an end-to-end trainable model that sequentially estimated 2.5D sketches and 3D object shape. Groueix${et}$ ${al.}$ \cite{groueix2018papier} used small patches to splice the result of 3D shapes. However, it is easy to cause overlapping between the small patches. Recently, some work\cite{xu2019disn,mescheder2019occupancy,chen2019learning} implicitly represented the 3D surface as the continuous decision boundary of a deep neural network classifier. In other words, these work learned a classifier to predict whether a point is inside or outside of the boundary, using this classifier as the shape representation. However, since the shape is represented by the weights of the classifier or regression model, these methods ignore some low-level shape information.

In this paper, we propose a multi-branch information fusion network for 3D reconstruction. We first strengthen the model's ability to capture the complex topology. To be more specific, as the difference between two different low-pass filtered images, the DoG is actually a band-pass filter, which removes high frequency components representing noise, and also some low frequency components representing the homogeneous areas in the image. The frequency components in the passing band are assumed to be associated to the edges in the images. So, we utilize DoG to  extract edge geometry and corners information, and use a separate branch network to infer the 3D shape. In addition, previous work generally used synthetic data for training, but suffered from domain adaptation when tested on real data. So, inspired by Li ${et}$ ${al.}$ \cite{li2020dynamic} and Xu ${et}$ ${al.}$ \cite{xu20203d}, we design several side branches at different locations of the main branch network to improve the generalization ability. Finally, we dynamically fuse the multi-branch probability distribution information as the final prediction result. As shown in \textbf{Fig.1}, we show the reconstruction results of our method on real images. Our main contributions lie in the following aspects:
\begin{itemize}
\item   We use DoG to process input images to extract edge geometry and corners information. Because we realize the object edge geometry and corners information are important for neural network to capture complex topology.
\item 	We design side branches from the intermediate layers of our neural network, so each side branch produces more diverse representations along its own pathway.
\item 	Unlike previous methods computing the average value or fixed weight of all branches predicted probability, we dynamically fuse the predicted probability of all branches to obtain the final predicted probability.
\item 	Extensive evaluation on a large-scale publicly available dataset ShapeNet demonstrates our method can achieve higher evaluation results than the state-of-the-art methods.
\end{itemize}

\section{Method}

\subsection{Overview}
Our goal is to use 2D image $\mathcal{X}$ and points $\mathcal{P}\in \mathcal{R}^3$ to infer the occupation probability of the corresponding points $\mathcal{P}$, i.e., to learn a mapping function$f_\theta: \mathcal{R}^3 \times \mathcal{X} \to [0,1]$. For a closed shape  $\mathcal{S}$, the binary neural network is equivalent to giving an occupancy probability between 0 and 1 for each point $\mathcal{P}_i$ to determine whether the point $\mathcal{P}_i$ is within the closed shape:
\begin{equation} \label{eqn1}
    \mathcal{S}(\mathcal{P}_i) =
    \begin{cases} 
    0  & \mbox{if }\mathcal{P}_i\notin shape \\
    1, & \mbox{if }\mathcal{P}_i\in shape  
    \end{cases}
\end{equation}

However, most of the existing work is only trained on synthetic data, and it is difficult to accurately predict when the model is tested on real data, i.e., the model lacks generalization ability. Secondly, it is difficult to recover the 3D shape accurately when the object has complex topological structure. To this end, we delicately design branches I,II,III from the intermediate layers of the main branch to assist inference. Finally, we dynamically fuse the prediction results of the main branch and the side branches. Therefore, first, we not only retain representation of the main branch but also generate more diverse representations along pathway of side branches. Second, we also use the DoG map to enhance the ability of our model to capture and learn the edge geometry and corners features information.

In order to better recover the complex topology and detailed information of the object and improve the generalization ability of the model. We design a multi-branch network to implement this idea. The workflow of DmifNet is shown in \textbf{Fig.2}. Specifically, we first utilize DoG to process the input image to obtain a Gaussian difference map. Second, the main branch infers the occupancy probability from the input 2D image. Third, the side branches I and II process the input information along its own pathway to infer the occupancy probability. Then, the side branch III uses the Gaussian difference map and the 2D image as input information to reason the occupancy probability. Finally, according to the different input images, we dynamically fuse the multi-branch probability information as the final prediction result.
% $$p(y|x,p)=\sum_{i=1}^B $$

\subsection{Main Branch and Side Branch I,II}
In this part, we make the main branch focus on learning the shape prior that explains input well. However, we observe that using a single mapping function is hard to learn the shape prior very well. So, we employ multiple mapping functions to more excellently and effectively learn the shape prior. As shown in \textbf{Fig.2}, we not only retain representative representation of the main branch but also generate more diverse representations along pathway of side branches.

\subsection{Side Branch III}
In this part, we use the DoG to process input image, which removes high frequency components representing noise, and also some low frequency components representing the homogeneous areas in the image. So, we use DoG maps as input information to enhance the model's ability to learn and capture the edge geometry and corners feature information. Then, our model can better retrieve the complex topology and detailed information of the object. In the first step, we convert the input RGB image into a gray image. Then, we convolve the gray image with different Gaussian kernels to obtain blurred images at different Gaussian scales:
\begin{align}
    F_i(x,y)&=G_{\sigma_i}(x,y)f(x,y)  \\ 
    & =f(x,y)\frac{1}{{\sigma_i}^d{2\pi}^{d/2}}exp-\frac{x^2+y^2}{2\sigma_i^2}\notag
\end{align}
where $f(x,y)$ represents the input gray image, $G_{\sigma_i}$ is the standard deviation of $i_{th}$ Gaussian kernel, $F_i(x,y)$ is  the output result of the input image after the $i_{th}$ Gaussian kernel convolution, $d$ is the dimension of the output. The second step, we use the subtraction of two adjacent Gaussian scale feature maps to obtain a Gaussian difference map:
\begin{align} 
    Diff&=F_{i+1}(x,y)-F_{i}(x,y)  \\
    &=(G_{\sigma_{i+1}}(x,y)-G_{\sigma_{i}}(x,y))f(x,y) \notag \\ 
    &=\frac{1}{{2\pi}^{\frac{d}{2}}}(\frac{1}{\sigma_{i+1}^d}exp(-\frac{r^2}{2\sigma_{i+1}^2})-\frac{1}{\sigma_{i}^d}exp(-\frac{r^2}{2\sigma_{i}^2}))f(x,y)\notag
\end{align}
 Note that the map keeps the spatial information contained in the frequency bands held between the two images. Where $r^2$ represents the blur radius. In addition, we find that the high-frequency random noise is removed by the DOG algorithm when tested on real data. In the third step, we concatenate the original input image and Gaussian difference map as input information to reason 3D shape. The whole process is shown in \textbf{Fig.3}
 
 \begin{figure}[htbp]
\centerline{\includegraphics[width=8.5cm]{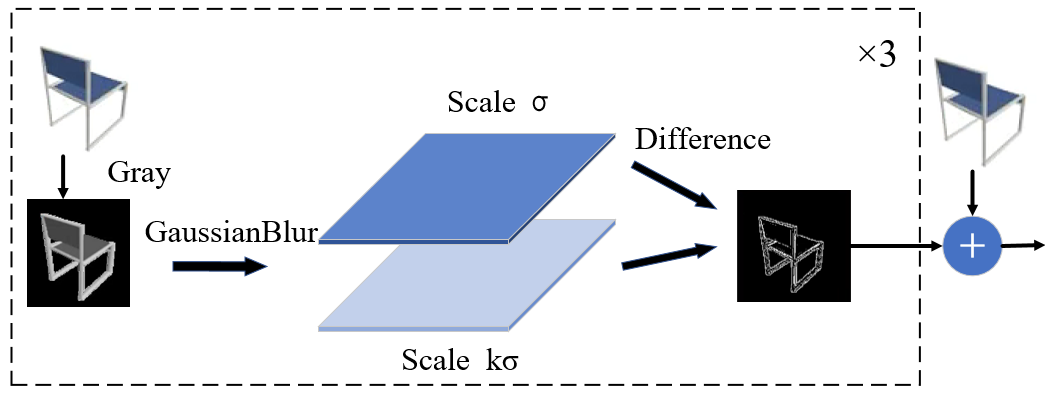}}
\caption{The process of side branch III preprocessing input image.  }
\label{fig}
\end{figure}

\subsection{Probability Mixture}
In this part, we model a linear combination of multiple branch networks to get a strong regressor. Specifically, when processing samples, the probability fusion module mixes the prediction results of each branch according to its contribution of each branch. Therefore, for each sample, all the branches will be integrated to infer the most accurate prediction probability:
\begin{align} 
  p(x)&=\sum_{i=1}^4 \alpha_i*\phi_i(x)
\end{align}
\[{s.t. \quad \alpha_i=f_\theta(\phi_i(x))}\]
where $\alpha_i$ represents the weight of each branch in the current sample prediction, $\phi_i(x)$ is prediction probability of each branch, $p(x)$ represents the output of the mixed prediction probability. $f_\theta(.)$ is the learned mapping function.

\subsection{Loss Function}
We train our network on the fully-annotated dataset ShapeNet\cite{chang2015shapenet}:
\begin{align} 
  Loss&=\frac{1}{|B|}\sum_{i=1}^{|B|}\sum_{j=1}^{K}L_{M_{CE}}(f_\theta(p_{ij},x_i),o_{ij})\\&+\frac{1}{|B|}\sum_{n=1}^{N}\sum_{i=1}^{|B|}\sum_{j=1}^{K}L_{S_{CE}}(f_\Theta{}_n(p_{ij},x_i),o_{ij})\notag
\end{align}

where $Loss$ represents the total training loss of a small batch $B$. $L_{M_{CE}}$ represents the main branch cross-entropy classification loss. $L_{S_{CE}}$ represents the side branch cross-entropy classification loss. $n=1,...,N$ represents the $n_{th}$ side branch. $f_\theta$ represents the main branch network parameters. $f_\Theta{}_n$ represents the $n_{th}$ side branch network parameter. The ${x_i}$ is the $i_{th}$ observation of batch B. $p_{ij}$represents the $j_{th}$ point of the $i_{th}$ observation. $o_{ij}$ is the ground truth of $p_{ij}$.

\subsection{Multi-branch Consistent Optimization}
In general, the optimization direction of the objective function of each branch in the multi-branch network is different. At the same time, the different optimization directions will negatively affect the accuracy of the model. So, we use a consistent optimization goal to optimize the multi-branch network. By the objective function equation (5), we not only directly collect the classification loss of each branch to optimize the network, but also pay attention to the different representations of each branch in its pathway. Once the knowledge is generated from the side branches or main branch, the network will achieve real-time knowledge interaction and information sharing through the common path between branches:

\begin{align} 
  Loss&=\frac{1}{|B|}\sum_{i=1}^{|B|}\sum_{j=1}^{K}L_{M_{CE}}(f_{\theta;I_{\Theta_n}}(p_{ij},x_i),o_{ij})\\&+\frac{1}{|B|}\sum_{n=1}^{N}\sum_{i=1}^{|B|}\sum_{j=1}^{K}L_{S_{CE}}(f_{_\Theta{}_n;I_{\theta}}(p_{ij},x_i),o_{ij})\notag
\end{align}
where $I_{\theta},I_{\Theta{}_n}$ represent the common path network parameters of each branch network.
% \begin{align} 
%   I_{\theta(l)}=
%      \begin{cases} 
%         0  & \mbox{if }l \notin \theta \\
%         1, & \mbox{if }l \in \theta  
%      \end{cases}
% \end{align}

\begin{table*}[t]
\begin{center}
\caption{ \textbf{Single Image 3D Reconstruction Results on ShapeNet}. Quantitative evaluations on the ShapeNet under IoU, Normal consistency and Chamfer distance. We observe that our method approach outperforms other state-of-the-art learning based methods in Normal consistency and IoU.}\label{tab:cap}. \label{tab:cap}
\setlength{\tabcolsep}{0.80mm}{
\begin{tabular}{ccccccccccccccccc} 
  \hline
  \textbf{IoU $\uparrow$} & Airplane & Bench & Cabinet & Car & Chair & Display & Lamp & Loudspeaker & Rifle & Sofa & Table & Telephone & Vessel  & Mean
  \\
  \hline
  3D-R2N2 \cite{choy20163d} ECCV'16       & 0.426 & 0.373 & 0.667 & 0.661 & 0.439 & 0.440 & 0.281 & 0.611 & 0.375 & 0.626 &0.420 &0.6118 &0.482  & 0.493  \\
  Pix2Mesh  \cite{wang2018pixel2mesh} ECCV'18  & 0.420 & 0.323 & 0.664 & 0.552 & 0.396 & 0.490 & 0.323 & 0.599 & 0.402& 0.613 & 0.395 & 0.661 &0.397 & 0.480 \\
   AtlasNet  \cite{groueix2018papier} CVPR'18  & - & - & - & - & - & - & -& - & - & - & - & - &- &-  \\
  ONet  \cite{mescheder2019occupancy} CVPR'19  & 0.571 & 0.485 & 0.733 & 0.737 & 0.501 & 0.471 & 0.371& 0.647 & 0.474 & 0.680 & 0.506 & 0.720 &0.530 &0.571  \\
 \textbf{Our}   & \textbf{0.603} & \textbf{0.512}& \textbf{0.753}&\textbf{0.758}& \textbf{0.542}& \textbf{0.560} & \textbf{0.416}& \textbf{0.675} & \textbf{0.493} & \textbf{0.701}& \textbf{0.550} & \textbf{0.750}& \textbf{0.574} &\textbf{0.607}  \\
  \hline
  \textbf{Normal Consistency $\uparrow$} & Airplane & Bench & Cabinet & Car & Chair & Display & Lamp & Loudspeaker & Rifle & Sofa & Table & Telephone & Vessel  & Mean\\
  \hline
   3D-R2N2 \cite{choy20163d} ECCV'16       & 0.629 &0.678& 0.782 & 0.714 & 0.663 & 0.720 & 0.560 & 0.711 & 0.670 &0.731&0.732 &0.817 &0.629  & 0.695  \\
  Pix2Mesh  \cite{wang2018pixel2mesh} ECCV'18  & 0.759 & 0.732 & 0.834 & 0.756 & 0.746 & 0.830 & 0.666 & 0.782 & 0.718& 0.820 & 0.784 & 0.907 &0.699& 0.772 \\
   AtlasNet  \cite{groueix2018papier} CVPR'18  & 0.836 & 0.779 & 0.850 & 0.836 & 0.791 & 0.858 & 0.694& 0.825 & 0.725 & 0.840 & 0.832 & 0.923 &0.756 &0.811  \\
  ONet  \cite{mescheder2019occupancy} CVPR'19  & 0.840 & 0.813 & 0.879 & 0.852 & 0.823 & 0.854 & 0.731& 0.832 & 0.766 & 0.863 & 0.858 & 0.935 &0.794 &0.834  \\
 \textbf{Our}   & \textbf{0.853} & \textbf{0.821}& \textbf{0.885}&\textbf{0.857}& \textbf{0.835}& \textbf{0.872} & \textbf{0.758}& \textbf{0.847} & \textbf{0.781} & \textbf{0.873}& \textbf{0.868} & \textbf{0.936}& \textbf{0.808} &\textbf{0.846}  \\
  \hline
  \textbf{Chamfer-$L_1$ $\downarrow$} & Airplane & Bench & Cabinet & Car & Chair & Display & Lamp & Loudspeaker & Rifle & Sofa & Table & Telephone & Vessel  & Mean\\
  \hline
   3D-R2N2 \cite{choy20163d} ECCV'16       & 0.227 &0.194& 0.217 & 0.213 & 0.270 & 0.314 & 0.778 & 0.318 & 0.183 &0.229&0.239 &0.195 &0.238  & 0.278  \\
  Pix2Mesh  \cite{wang2018pixel2mesh} ECCV'18  & 0.187 & 0.201 & 0.196 & 0.180 & 0.265 & 0.239 & 0.308 & 0.285 & 0.164& 0.212 & 0.218 & 0.149 &0.212& 0.216 \\
   AtlasNet  \cite{groueix2018papier} CVPR'18   & \textbf{0.104} & \textbf{0.138}& 0.175&\textbf{0.141}& 0.209& \textbf{0.198} & \textbf{0.305}& \textbf{0.245} & \textbf{0.115} & \textbf{0.177}& 0.190 & 0.128& \textbf{0.151} &\textbf{0.175}  \\
  ONet  \cite{mescheder2019occupancy} CVPR'19  & 0.147 & 0.155 & 0.167 & 0.159 & 0.228 & 0.278 & 0.479& 0.300 & 0.141 & 0.194 & 0.189 & 0.140 &0.218 &0.215  \\
 \textbf{Our}   & 0.131 & 0.141 &\textbf{0.149} & 0.142 &\textbf{0.203} & 0.220 & 0.351& 0.263 & 0.135 & 0.181 & \textbf{0.173}  &\textbf{0.124} &0.189 &0.185  \\
  \hline
  \multicolumn{4}{l}{$^{\mathrm{a}}$ The Bold-faced numbers represent the best results.}
\end{tabular}}
\end{center}
\end{table*}

\section{Experiment}
In this section, we first introduce settings and implementation details for training and evaluation. Second, we report our results which are evaluated on 13 categories from the ShapeNet repository and compare the performance of our method to several state-of-the-art methods. Finally, we use ablation study to validate each module in our model.
\subsection{Dataset}
 ShapeNet\cite{chang2015shapenet}: ShapeNet is a richly-annotated, large-scale dataset consisting of 50000 models and 13 major categories. We split the dataset into training and testing sets, with 4/5 for training and the remaining 1/5 for testing. 

Online Products\cite{oh2016deep}: The dataset contains images of 23,000 items sold online.  Since the dataset does not have the ground-truth, we only use the dataset for qualitative evaluation.

\subsection{Metrics}
Metrics: Following the work\cite{michalkiewicz2020simple} experimental setup, we use the volumetric Intersection over Union (IoU), the Normal consistency score (NC), and the Chamfer-L1 distance (CD) to evaluate our method. 
The first one is Intersection over Union (IoU) between prediction $\mathcal{R}$ and ground truth shape $\mathcal{G}$:

\begin{align} 
  IoU(\mathcal{R},\mathcal{G})&=\frac{|\mathcal{R} \bigcap \mathcal{G}|}{|\mathcal{R} \bigcup \mathcal{G}|}
\end{align}

The normal consistency (NC) between the normals in prediction generated mesh $\mathcal{R}$  and the normals at the corresponding nearest neighbors in the ground truth generated mesh $\mathcal{G}$  is defined as:

\begin{align} 
  NC(\mathcal{R},\mathcal{G})&=\frac{1}{|\mathcal{R}|} \sum_{r\in \mathcal{R}}^{g \in \mathcal{G}}|r\cdot g|+\frac{1}{|\mathcal{G}|} \sum_{g\in \mathcal{G}}^{r \in \mathcal{R}}|g\cdot r|
\end{align}

The Chamfer distance (CD) between the ground truth $\mathcal{G}$ and the predicted shape $\mathcal{R}$  is defined as:
\begin{align} 
  CD(\mathcal{R},\mathcal{G})&=\frac{1}{|\mathcal{R}|} \sum_{r\in \mathcal{R}} min_{g \in \mathcal{G}}||r-g||_2\\&+\frac{1}{|\mathcal{G}|} \sum_{g\in \mathcal{G}} min_{r \in \mathcal{R}}||g-r||_2\notag
\end{align}

\subsection{Implementation Detail}
We train our network using Adam\cite{kingma2014adam}, adopting a starting learning rate of 0.004. We implement our code using python 3.6 on Pytorch 1.0.0, which trains on the ShapeNet dataset in one Nvdia Titan Xp GPU with CUDA 9.0 and cudnn7. Follow the work\cite{wu20153d}, we employ a ResNet18 architecture as encoder, then we use a fully-connected neural network with 5 ResNet blocks \cite{he2016deep} as decoder and condition it on the input using conditional batch normalization (CBN) \cite{de2017modulating,dumoulin2016adversarially}. As shown in \textbf{Fig.4}.

\begin{figure}[htbp]
\centerline{\includegraphics[width=8.5cm]{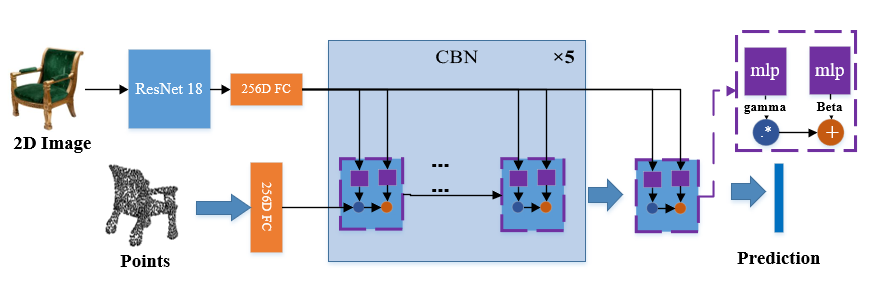}}
\caption{The architecture of encoder and decoder network. Best viewed in color.}
\label{fig}
\end{figure}

\begin{figure}[htbp]
\centerline{\includegraphics[width=8.5cm]{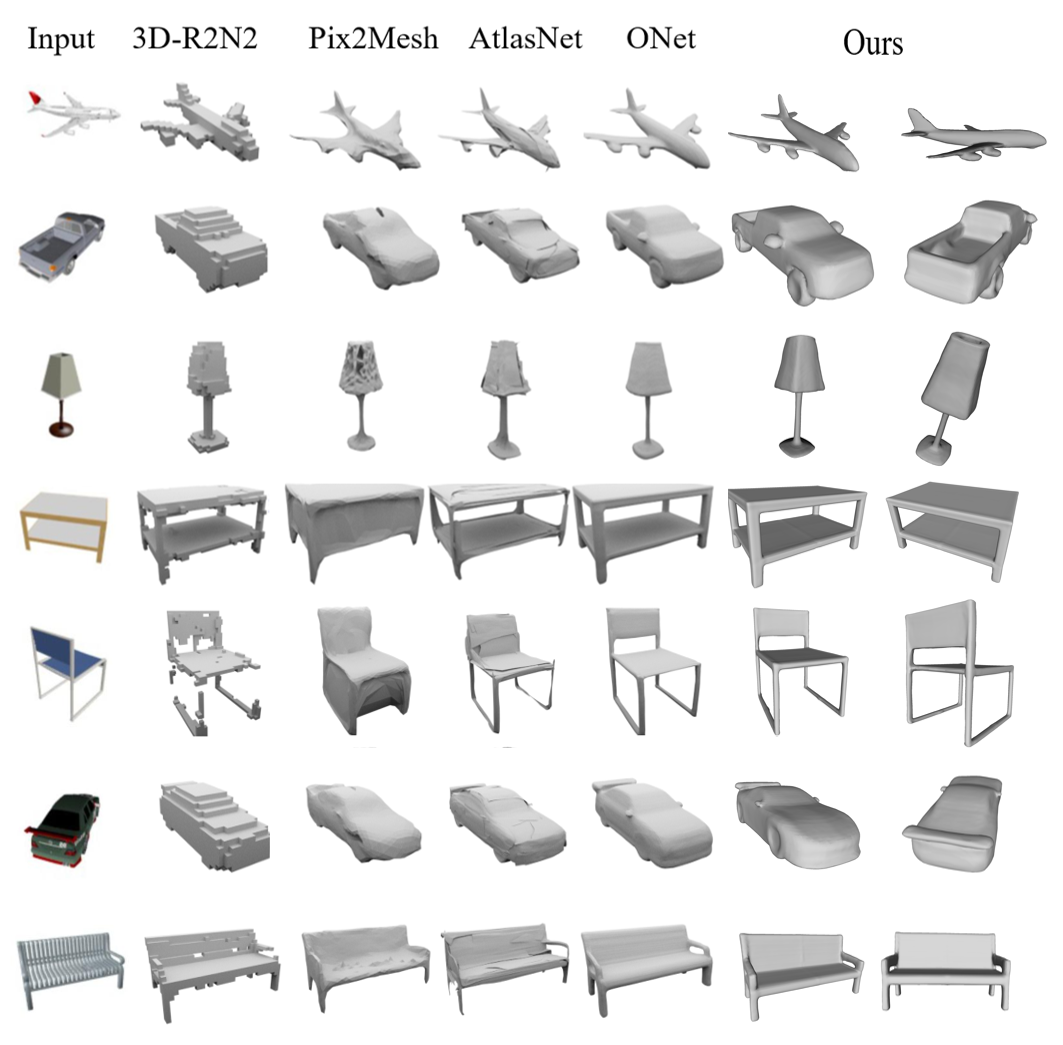}}
\caption{Single Image 3D Reconstruction on ShapeNet. The first column is the input 2D images, the last two columns are the results of our method. The other columns show the results for various methods. Best viewed on screen with zooming in.}
\label{fig}
\end{figure}

\subsection{Quantitative Results on ShapeNet}
Quantitative results on ShapeNet are given in \textbf{TABLE I}. For the IoU metric, we observe that our method outperforms other state-of-the-art methods in all categories of ShapeNet. Compared to other methods, our method gives significant promotion in the Normal consistency score metric than all works. For the Chamfer-L1 distance metric, while not trained with regard to Chamfer distance as Pixel2Mesh and AtlasNet, we also observe that our method surpasses other methods except Groueix ${et}$ ${al.}$ \cite{groueix2018papier}.

\subsection{Qualitative Results on ShapeNet}
In \textbf{Fig.5}, we evaluate the performance on the single-view image reconstruction qualitatively. We observe that all methods can capture the basic geometric information of the input image. However, 3D-R2N2 lacks lots of details on complex topology objects. In contrast, Pix2Mesh is able to better capture geometric information, but objects with more complex topologies have deformations and holes. Similarly, AtlasNet can capture geometric information well, but AtlasNet is easy to produce self-intersections and overlapping since it is assembled surfaces from small patches. Moreover, ONet is able to capture geometric information quite well and produce a high-fidelity output, but lacks some details in edges and corners. Finally, our method is able to capture complex topologies, produce high-fidelity 3D shape and preserve most of edges and corners details.

\begin{table}[htbp]
\caption{ \textbf{Ablation Study.} when we ablate our model step by step, we observe that the performance of the model degrades. First, we use B0 as final model to test on ShapeNet. Second, we use B0 with B1 and B2 as final model to test on ShapeNet. Third, we use B0 with B1, B2, and PMM as final model. Finally, we use B0 with B1, B2, B3, and PMM as final model.}
\begin{center}
\begin{tabular}{|c|c|c|c|}
\hline
\textbf{ Test }&\multicolumn{3}{|c|}{\textbf{Metrics }} \\
\cline{2-4} 
\textbf{Model} & \textbf{\textit{IoU $\uparrow$}}& \textbf{\textit{ NC $\uparrow$ }}& \textbf{\textit{ Chamfer $\downarrow$}} \\
\hline
Model w/o $B_3$,PMM,$B_2$and$B_1$&  0.593&0.840 & 0.194 \\
Model w/o $B_3$,PMM&  0.602&0.842 & 0.191 \\
Model w/o $B_3$&   0.604&0.842 & 0.185 \\
Full model&  \textbf{0.607}&\textbf{0.846} & \textbf{0.185} \\

\hline
\multicolumn{4}{l}{$^{\mathrm{a}}$ The Bold-faced numbers represent the best results.}
\end{tabular}
\label{tab1}
\end{center}
\end{table}

\subsection{Ablation Study}
We calculate the ablation study on the proposed method, mainly focus on the effects of the side branches and the probability mixture module. We denote the main branch as $B_0$, and side branches I, II are denoted as $B_1$ and $B_2$ respectively. Then, side branch III and probability mixture module are denoted as $B3$ and PMM respectively. The results are shown in \textbf{TABLE II}.

We first examine how $B_1$ and $B_2$ affect the performance of our model. $B_1$ and $B_2$ considering the generalization ability of the model, we utilize them to produce more diverse representations along its own pathway. So, we observe that the $B_0$ and $B_1$ branches perform well on  metrics of IoU and NC.

To test the effect of the  components of PMM, we utilize PMM to dynamically fuse prediction probabilities. Similarly, we observe that the PMM effectively promotes the metrics of IoU and Chamfer-$L_1$.

Finally, we discuss the effect of $B_3$ on model performance. We employ $B_3$ to enhance the edge geometry  and corners information. As shown in Table 2, we observe that the $B_3$ significantly improves the metrics of IoU and NC.

\begin{figure}[htbp]
\centerline{\includegraphics[width=8.5cm]{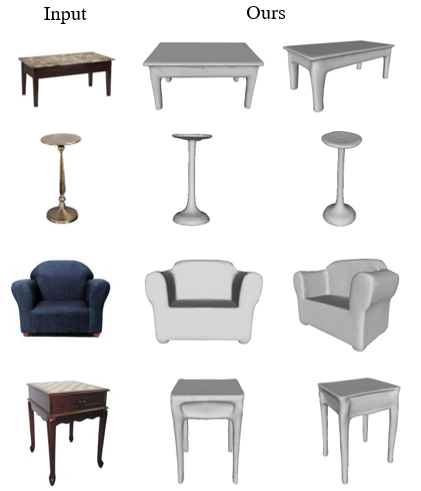}}
\caption{Single Image 3D Reconstruction on Real Data. The first column is the input 2D images, the other columns are the reconstructed results of our method in different viewpoints. Best viewed on screen with zooming in.}
\label{fig}
\end{figure}

\subsection{Qualitative Results on Real Data}
In order to test the generalization ability of our network on real data, we apply our model to the Online Products datasets\cite{oh2016deep} for qualitative evaluation. Note that our network is not trained on this dataset.

Several qualitative results are shown in \textbf{Fig.6}. We carefully select some representative images in the dataset to display the qualitative results. We give the reconstruction results under two viewpoints. Through the results, we observe that although our model is only trained on synthetic data, but it also has good generalization for real data.

\section{Conclusions}

In this paper, we propose a novel approach based on dynamic multi-branch information fusion for 3D reconstruction. Our method has addressed the problem of complex topology and detailed information, which are difficult to recover accurately. We introduce side branches from the intermediate layers of our neural network to make it produce more diverse representations. By utilizing DoG to enhance edge geometry information and  mixing multiple branches reasonably, our method is competent for handling the single-view image reconstruction task. Extensive experiments demonstrate our method can boost the model generalization ability and recover the complex topology and detail-rich 3D shape from a 2D image. Since a large number of 3D datasets do not have the ground-truth, so previous work only used these datasets for qualitative evaluation. Inspired by some latest work\cite{xie2020self,kingma2014semi,zhu2009introduction}, in the future work, we plan to introduce  semi-supervised learning in order to take advantage of these 3D datasets reasonably.

\bibliographystyle{IEEEbib}
\bibliography{dmifnet}
\end{document}